\documentclass[letterpaper, 10 pt, conference]{ieeeconf}  

\IEEEoverridecommandlockouts                              

\usepackage{graphics} 
\usepackage{epsfig} 
\usepackage{mathptmx} 
\usepackage{times} 
\usepackage{amsmath} 
\usepackage{amssymb}  
\usepackage{amsfonts}
\usepackage{algorithmic}
\usepackage{graphicx}
\usepackage{textcomp}
\usepackage{xcolor}
\usepackage{hyperref}

\title{\LARGE \bf
PLUME: Procedural Layer Underground Modeling Engine
}

\author{Gabriel Manuel Garcia$^{1}$, Antoine Richard$^{1}$ and Miguel Olivares-Mendez$^{1}$
\thanks{*This research was funded in whole, or in part, by the Luxembourg National Research Fund (FNR), grant reference [CAESAR-XR/Gabriel/17679211]. For the purpose of open access, and in fulfilment of the obligations arising from the grant agreement, the author has applied a Creative Commons Attribution 4.0 International (CC BY 4.0) license to any Author Accepted Manuscript version arising from this submission.}
\thanks{$^{1}$Affiliated at  Space Robotics (SpaceR) Research Group, Interdisciplinary Centre for Security, Reliability and Trust (SnT), University of Luxembourg, Luxembourg, Luxembourg {\tt \{gabriel.garcia}, {\tt antoine.richard}, {\tt miguel.olivaresmendez\}@uni.lu}}
}

\begin{document}

\maketitle
\thispagestyle{empty}
\pagestyle{empty}

\begin{abstract}
As space exploration advances, underground environments are becoming increasingly attractive due to their potential to provide shelter, easier access to resources, and enhanced scientific opportunities. Although such environments exist on Earth, they are often not easily accessible and do not accurately represent the diversity of underground environments found throughout the solar system.

This paper presents PLUME, a procedural generation framework aimed at easily creating 3D underground environments. Its flexible structure allows for the continuous enhancement of various underground features, aligning with our expanding understanding of the solar system.

The environments generated using PLUME can be used for AI training, evaluating robotics algorithms, 3D rendering, and facilitating rapid iteration on developed exploration algorithms. In this paper, it is demonstrated that PLUME has been used along with a robotic simulator.
PLUME is open source and has been released on Github. \href{https://github.com/Gabryss/P.L.U.M.E}{https://github.com/Gabryss/P.L.U.M.E}
\end{abstract}

\section{Introduction}



To do planetary exploration, shelter will be an essential part to keep robots and humans protected from extreme temperatures, solar radiation, and micrometeorites. Existing subsurface structures such as caves or lava tubes are considered of high interest to create shelter. In addition to the provided shelter, these environments offer valuable scientific data by providing insight into the internal composition of celestial bodies (e.g., geology for Earth, selenology for the Moon, or areology for Mars), eliminating the need for ground excavation and the associated resource requirements. Such environments could also benefit companies looking to harvest resources for in situ processing and utilisation, further emphasising the importance of getting used to these environments.

To prepare our robotic platforms for cave underground environments, Earth's underground environment can be leveraged to overcome potential difficulties, as demonstrated by the DARPA Subterranean Challenge \cite{c_intro1}, where multiple teams competed in an underground environment while tackling various challenges. Although these environments are fascinating, they do not represent the full diversity of caves on Earth, let alone those on other celestial bodies. 

\begin{figure}[htbp]
\centerline{\includegraphics[width=1.0\linewidth]{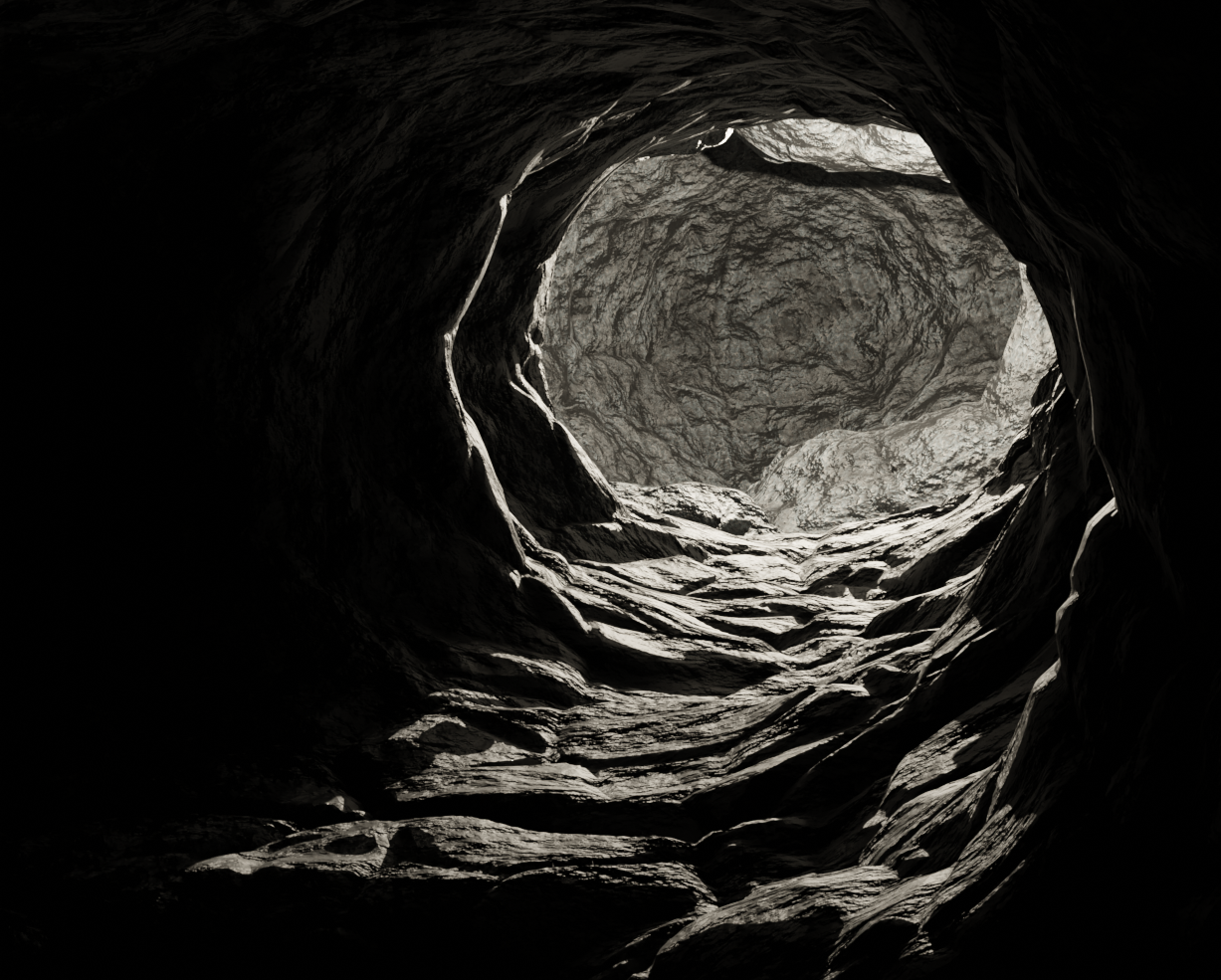}}
\caption{Procedural generated cave with PLUME}
\label{fig:cave_render}
\end{figure}


To address this challenge, we present a procedural generator with a new approach called PLUME: Procedural Layer Underground Modeling Engine.


This procedural generation approach offers several benefits, including the ability to provide virtual environments that are easily and rapidly created, thereby mitigating the logistical limitations and difficulties of the real environment. Furthermore, PLUME offer the possibility to run multiple missions in parallel or in a short amount of time. Procedurally generated simulation environments provide a large variety of environments, without the constrains of elaborate logistical work to access a different sites for experiments. PLUME also offer the possibility to try a more aggressive algorithm without damaging the platform or acquiring expensive sensors, in fact, simulations are scalable and cost effective with little overhead and preparation time, once they are set up for the target application. As artificial intelligence continues to develop, extensive training is required to achieve interesting behaviours in robots, particularly when it comes to various forms of reinforcement learning. In this regard, procedurally generated caves mitigate overfitting by providing a diverse range of easily accessible environments for training purposes.


PLUME aims to provide a simple and efficient framework for generating underground environments. It produces a complete mesh with baked textures in the desired format (obj, ply, usd, fbx for the mesh and png, jpg for the texture) using input parameters or a graph as an input. This generated environment can be used in various simulation software, including Gazebo, Isaac Sim, Unity, and Unreal Engine. In addition, the framework can be easily improved to incorporate more extensive underground structures, thereby increasing the variety of environments available to the research teams to work with.

\section{Related work}
Procedural generation of underground environments has been a subject of study for a long time. The first and most intuitive generation method for such environments was introduced with cellular automata, as found in \cite{c1}, \cite{c2}, and \cite{c3}. It uses a 2D grid with each cell either dead or alive. Then, an initial set of alive cells is placed onto the grid and evolves according to a set of rules. The user can decide to stop the simulation after a given amount of steps and the resulting map is our underground environment, with alive cells representing the cave and dead cells representing the rocks. This approach is well suited in 2D but has its limits in terms of computational time in 3D and produces less appealing natural results as the scale of the grid increases \cite{c4}. Other similar methods can be applied, such as Midpoint Placement, Random Walk, Perlin Worms, or hybrid methods between those, but they all share similar limitations to cellular automata.

Another intuitive approach would be to make a set of tiles and randomly place them on a 2D map. 
In \cite{c5}, a set of tiles (intersections, corridors and blocking tiles) models with varying numbers of connection points has been created. The tiles are then randomly placed on a 2D map by choosing one of the open connections for the newly added tile. This process entails a very low computational cost, making it possible to generate results in real time. However, due to a limited amount of 3D assets, it also has the drawback of being highly repetitive. 

Similarly to the 2D methods seen previously, but without the drawbacks of the repetitiveness of the tile assets, an initial heightmap selected from an existing region (Montana) can be used to get the "skeleton" of the cave as well as the cave profile by mirroring the heightmap as seen in \cite{c6}. This process is interesting as it takes existing data and generates a root-like cave system; however, it is limited to a flat representation and may not include caves with slopes or multiple levels.

As seen in \cite{c7}, extended with \cite{c8}, and \cite{c9}, voxel-based procedural generation can produce interesting results, as it directly generates the environment considering its surroundings. In the paper \cite{c9}, physics-based generation is also considered to create realistic underground environments. The downside of using a voxelised environment is the voxel resolution, which is directly linked with the quality of the representation of the environment. Increasing the resolution directly impacts the computational resources required for generating high-quality environments. This process makes it less suitable for real-time generation.

In order to minimise the computation time, \cite{c10} proposes to generate a cave system using the GPU instead of the CPU. This approach drastically speeds up the generation time of the environment while still using the voxel approach. On the other hand, the skeleton of the cave is not driven by a two-dimensional randomness algorithm or physics, but by a graph system called the L-system. 

The papers \cite{c10}, \cite{c11}, and \cite{c12} present the L-system as a collection of letters, each letter representing an instruction for a virtual turtle to follow in order to draw the graph (also referred to as the "skeleton") of the cave. It is important to note that the L-system approach can be applied in both continuous and discrete environments. In a continuous environment, a graph can be generated as demonstrated in \cite{c10}. In contrast, a discrete environment can incorporate grids, cellular automata, and the L-system, as illustrated in \cite{c11} and \cite{c12}.

To avoid repetitive patterns as much as possible, continuous generation with graph generation is investigated, as represented in the papers \cite{c13} and \cite{c14}. Those papers yield very good results in procedural generation using a continuous graph, while being open-source and providing a good representation of the underground. The paper \cite{c13} focuses on the physical procedural generation of the environment using a geology-based function to generate environments as close to the real karstic cave networks. On the other hand, the paper \cite{c14} focuses on generating tunnel-like networks with the ability to utilise them in robotic simulation software such as Gazebo. This robotic integrated pipeline is what we aim for with this paper.

As shown in \cite{c15} and \cite{c16}, there is a growing interest in robotic simulation and real field experiments of underground environments. This interest is particularly appreciated due to the challenging conditions in such an environment, which necessitate training and experimenting with algorithms in simulation before deploying them in the field. In terms of space, recent papers \cite{c17} and \cite{c18} demonstrate the potential benefits and interest in using lava tubes, also known as pyroducts, as a base shelter for further exploration.

\section{Methodology}
PLUME is an open source Blender-based pipeline that generates 3D meshes and high-quality textures of various types of underground environments. It is adequately modular to allow for easy extension with additional graph algorithms. The currently generated environment can take the form of a mine or a cave, but with a huge focus on the latter. 
\begin{figure*}[htp]
\centering
\includegraphics[width=1.6\columnwidth]{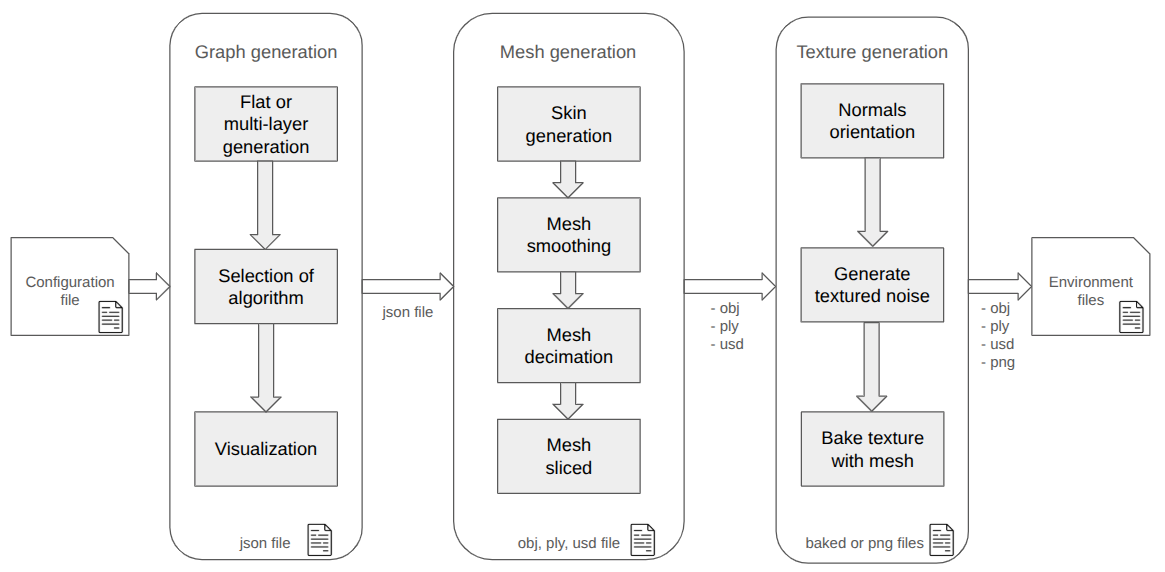}
\caption{Overview of the procedural generation process of PLUME}
\label{fig:plume_overview}
\end{figure*}

As seen in Fig. \ref{fig:plume_overview} PLUME is composed of three distinct stages: Graph generation, mesh generation and texture generation.

Each stage is composed of distinct blocks that represent the generation steps within a stage. However, some blocks are optional and can be skipped either to speed up the generation time or because they are not needed for the desired environment.

\subsection{Graph generation}
The first stage of the generation, as shown in Fig. \ref{fig:plume_overview}, is the graph generation. The graph represents the "skeleton" of the underground environment. Depending on the algorithm, the generated environment can take the form of a cave, a mine, or any other type of underground setting. The generation of the graph can either be single or multi-layer. A single layer graph is a graph generated on a plane with only one layer of depth. The multi-layer generation is a 3D graph generated with multiple layers (composed of single layer graphs) with various connections between them. The option to generate single or multi-layer graphs allows for various graph structures that do not require a multi-level underground environment.

This framework provides high flexibility for the graph generation. It allows users to use their graph as input (either from a previous generation or a handcrafted graph). The framework's structure is modular and enables users to interface with new algorithms to generate new graphs easily.

Fig. \ref{fig:plume_overview} provides a global overview of the framework's pipeline but does not cover the details and algorithms of each block, as they can be easily swapped to change the output environment. However, native graph generation is based on a modified Gaussian-Perlin algorithm to produce the best visual output, according to our assessment, and to improve the environment traversability for wheeled rovers. 

\begin{figure}[htbp]
\centerline{\includegraphics[width=0.9\linewidth]{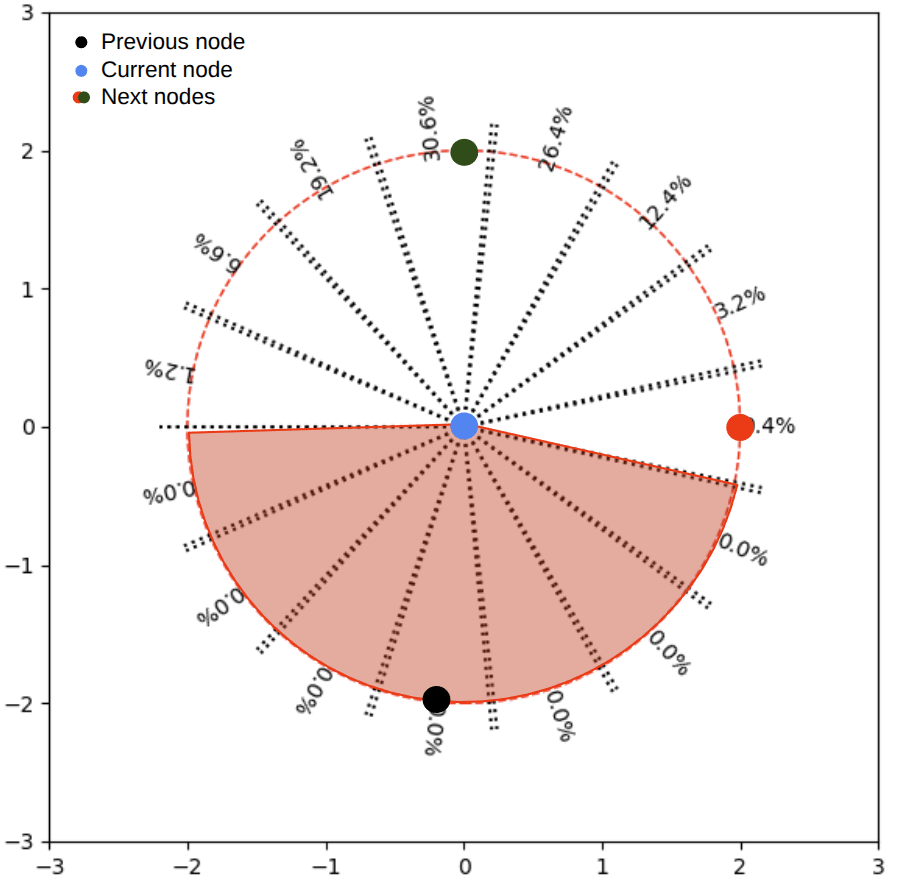}}
\caption{Representation of a node during the generation process. The red highlighted area represent the "forbiden" zone preventing the graph to circle on itself. Each dots represent a node. The percentage on the circle represent the node spawning probability distribution around the current node.}
\label{fig:plume_node}
\end{figure}

The graph is composed of a defined number of nodes specified in the configuration file, and each node has six distinct parameters: ID, parent, edges, coordinates, radius, and active. The active parameter indicates whether the node is still capable of generating new nodes or if this branch died during generation. This allows for random stopping of some branches to expand, whereas others can still grow in different directions.
The parent indicates which node created the current node, whereas the edges show the connection between the nodes. The coordinate parameter enables the representation of the node in a 3D environment for later visualisation and rendering. Finally, the radius indicates how far the next node is created by making a circle around the current node. New nodes can only be created on a circle around the current node; this particular representation can be seen in the Fig. \ref{fig:plume_node}.

At the base of native graph generation, the algorithm starts with an origin node at the position $(x_0,y_0,z_0) = (0,0,0)$ and randomly creates nodes on its circle. Then, each new node makes its own circle given a random radius (within the range defined in the configuration file). The process of creating new nodes is shown in Fig. \ref{fig:plume_node} with a black dot representing the previous node, a blue dot representing the current node, and green/red dots representing new nodes. To prevent the graph from circling on itself, a forbidden zone is made on each circle along the direction of the previous node as shown in the Fig. \ref{fig:plume_node}. The randomisation of the nodes along the circle is taken care of by a Gaussian, Perlin or a hybrid distribution. 
The circle is divided into 360 sections that represent the probability percentage of creating a new node at a given angle. The sections of the forbidden area have a probability of creating a new node of $0\%$.
The sections are stored in a 1D list, and the probability distribution is applied to it.

\begin{figure}[htbp]
\centerline{\includegraphics[width=0.7\linewidth]{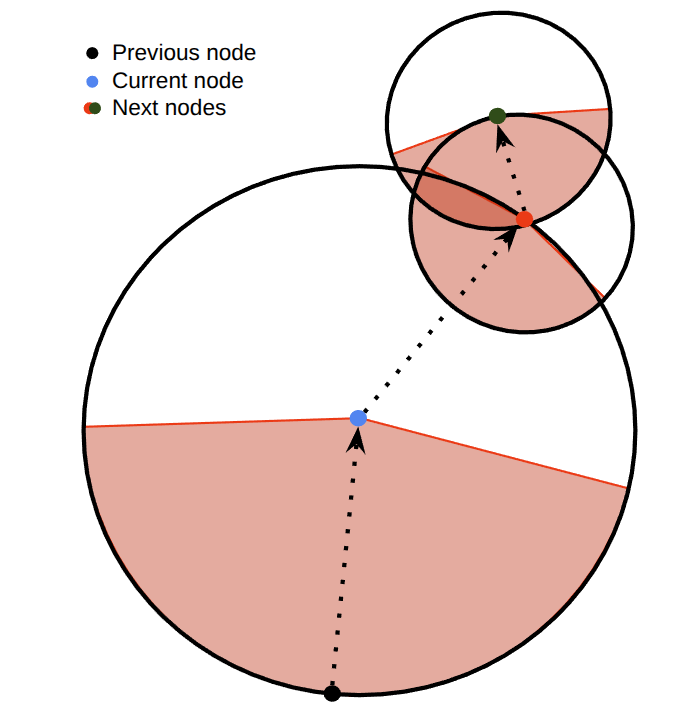}}
\caption{Graph generation process. Each dot represents a node, and each red area represents the forbidden node creation angle.}
\label{fig:plume_graph}
\end{figure}

Fig. \ref{fig:plume_graph} shows how the graph is created at each step of the generation process, leading to a tree-like structure. For simplicity, the figure shows only one node in each circle, but there can be multiple nodes based on the randomised values defined in the configuration file.

This generation method enables the rapid creation of graphs while producing visually appealing results in the underground environment. The radius of each node is used during the mesh generation process, making the entire environment determined by its graph data, which allows for reproducibility.

\begin{figure}[h]
\begin{tabular}{ll}
\includegraphics[scale=0.17]{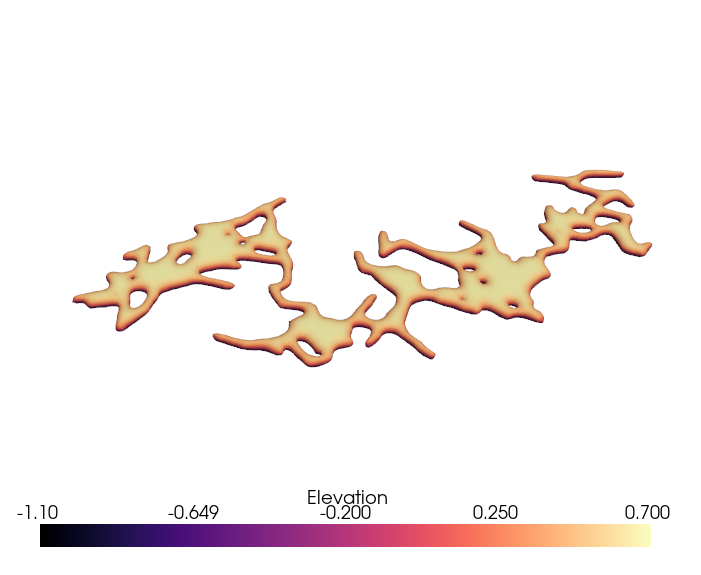}
&
\includegraphics[scale=0.165]{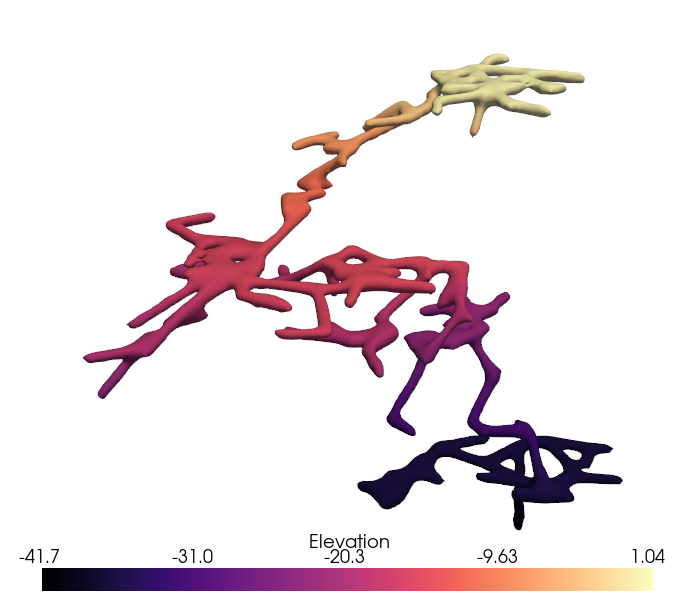}
\end{tabular}
\caption{Left: Single layer underground graph. Right: Multi-layer underground graph. The colour gradient illustrates the elevation of the environment.}
\label{Fig:plume_graph_representation}
\end{figure}

Fig. \ref{Fig:plume_graph_representation} shows the single and multi-layer graph generation of an underground environment. The single layer representation has a small local elevation to improve its realism. As shown in the figure, clusters of nodes create rooms of various sizes. They are interconnected through a network of passages, allowing the rover to explore different environments. The multi-layer network has its layers interconnected through multiple angles. The maximum interconnection angle can be defined in the configuration file. On the right of Fig. \ref{Fig:plume_graph_representation}, the upper layers are connected with a gentle slope, whereas the lower layers have a connection of almost 90\(^\circ\) angle between them.

In addition to the graph data, this stage outputs a \textit{JSON} file containing all the information provided in the configuration file, allowing later reproducibility with a different mesh or texture generation. This \textit{JSON} file can easily be shared to generate environments between teams rather than sharing a heavy environment file.

\subsection{Mesh generation}

To facilitate the generation of the environment, PLUME relies on the Blender API to create a high-resolution mesh and texture it later. Given all the data from the previous stage, Blender loads the graph and starts making the mesh around it. Blender works with geometry nodes that execute a mathematical function. In our case, we first create a skin around the graph to generate a tube-like environment. Before proceeding to the next step, some parts of the mesh may exhibit artefacts, resulting in an unrealistic appearance. To remove the artifacts, a smoothing geometry node is applied, but the resulting mesh has a very high amount of polygons and therefore needs to be decimated before the next step. The decimation process reduces the level of detail and, therefore, the computational load. This step can be fine-tuned in the configuration file.

\begin{figure}[h]
\begin{tabular}{ll}
\includegraphics[scale=0.11]{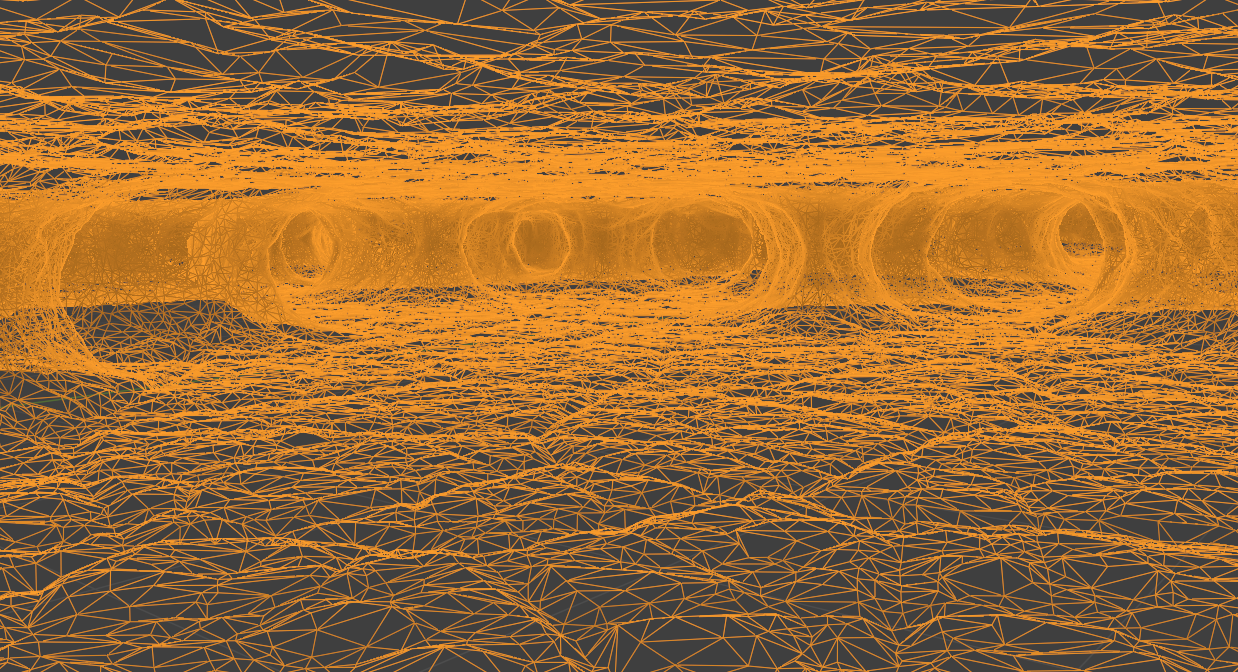}
&
\includegraphics[scale=0.10]{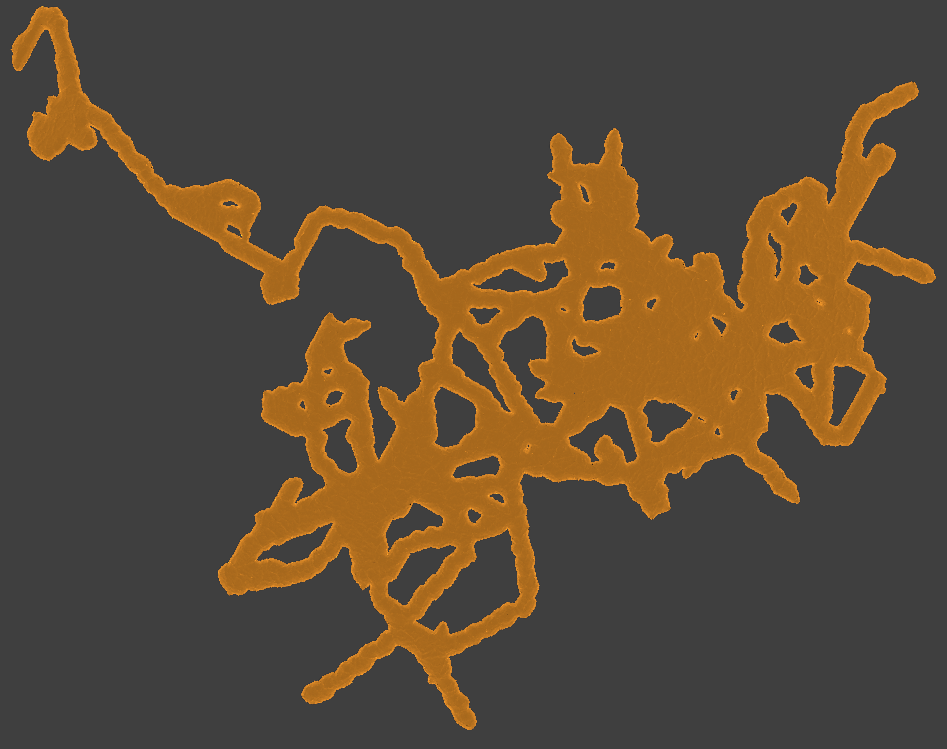}
\end{tabular}
\caption{Left: Single layer underground mesh. Right: Top-down view of a single layer mesh generation}
\label{Fig:plume_mesh_representation}
\end{figure}

Fig. \ref{Fig:plume_mesh_representation} shows the interior of a single layer underground mesh on the left, while on the right it shows the entire mesh generation. As seen in the left visualisation, the resulting mesh contains a large number of polygons, ensuring a fine level of detail in a simulation software.

PLUME initially generates one single mesh with a single texture file. If the mesh is very large, the resolution of the texture does not change, resulting in a low-quality texture. To avoid this limitation, the mesh is split into multiple chunks, and every chunk received its own texture. This separates the texture quality from the total size of the mesh and allows for consistent texture quality across different generations.

\begin{figure}[htbp]
\centerline{\includegraphics[width=0.7\linewidth]{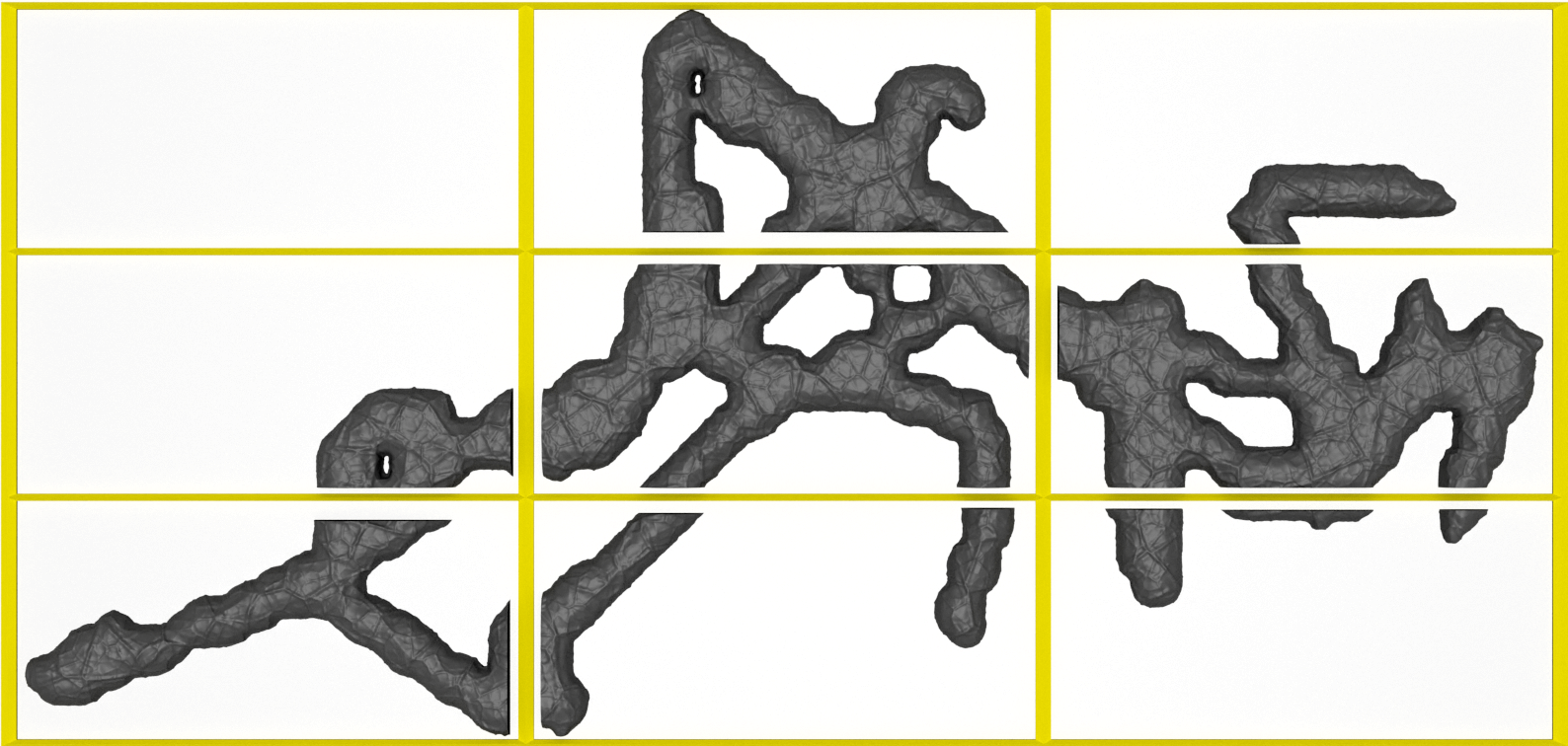}}
\caption{Single layer mesh generation sliced into smaller chunks (top-down view)}
\label{fig:cave_chunks}
\end{figure}

In Fig. \ref{fig:cave_chunks}, a grid is created along all three axes of the mesh, although the $z$ axis is not used for single layer generation. The mesh is divided into separate chunks. Each chunk contains a portion of the overall mesh along with its associated textures.

\subsection{Texture generation}

\begin{figure}[h]
\begin{center}
\begin{tabular}{lll}
\includegraphics[scale=0.14]{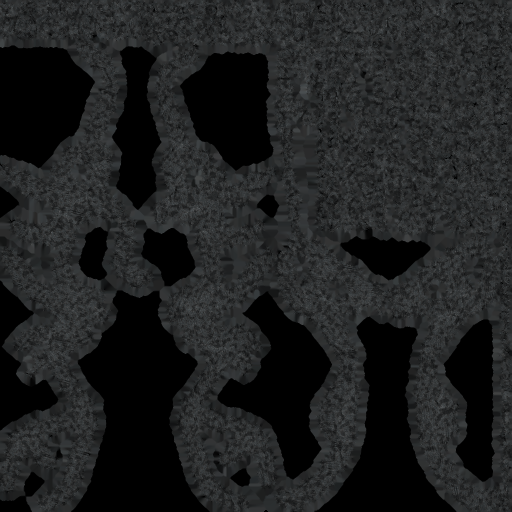}
&
\includegraphics[scale=0.14]{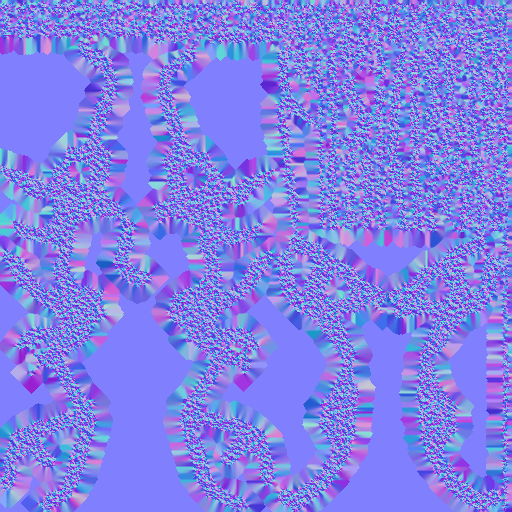}
&
\includegraphics[scale=0.14]{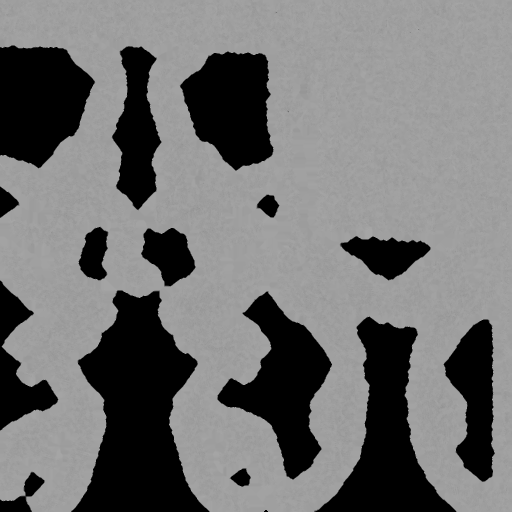}
\end{tabular}
\end{center}
\caption{From left to right: colour texture, normal texture and roughness texture}
\label{Fig:plume_texture}
\end{figure}

Once the mesh is generated and subdivided into multiple chunks, the texture can be baked and applied to the chunks. The texture generation was completed and applied before the mesh chunk division to maintain a coherent overall look. The texture generation is entirely procedural and is based on Voronoi and Perlin noise. As seen in Fig. \ref{Fig:plume_texture}, each chunk is composed of 3 different visual textures: the colour, the normal, and the roughness texture. The colour gives the overall visual appearance of the mesh. The normal texture gives the mesh a 3D aspect while saving computational cost, and the roughness represents the amount of light reflection on the surface. The roughness can be adjusted to make a cave more humid, and hence change its visual aspect for VSLAM (Visual Simultaneous Localisation and Mapping).


Once the mesh and textures are generated, they are baked together in order to map the textures with the mesh in a format that is usable in any simulation software. As seen in Fig. \ref{fig:cave_render} PLUME environment generation has a realistic look, which is very important for VSLAM that rely on cameras.

At each stage, the generation can be paused or resumed, and pre-visualisation enables the user to visualise the result without undergoing the lengthy and costly generation process.

\section{Results}

\subsection{Material used}
The specifications of the used computer include an Intel i5-12400F CPU, an Nvidia RTX 3070 GPU used for the texture baking, 32 GB of RAM, and a 1 TB NVMe SSD, with 32 GB allocated for swap memory if needed. The current PLUME pipeline use Blender version 4.0.1 and Cycles as a rendering engine. The speed of the generation depends heavily on the texture resolution and the number of chunks. The table \ref{tab:generation_time} shows different generation times given the computer used for various parameters in the configuration file.

\begin{table}[h!]
    \begin{center}
        \caption{Computation time to generate environment using PLUME}
        \label{tab:generation_time}
        \begin{tabular}{ l c c c c} 
         Parameters                   & 50N(1K)   & 50N(4K) & 250N(1K) & 250N(4K)\\ 
         \hline
         Single (no chunks)             & 00m22s      & 02m15s       & 01m02s     & 02m46s    \\ 
         Multi (no chunks)              & 00m31s      & 02m28s       & 01m42s     & 03m50s    \\ 
         Single (4 division)            & 07m05s      & 29m37s      & 10m01s    & 77m43s   \\ 
         Multi (4 division)             & 11m51s     & 126m49s     & 147m52s   & 219m49s  \\ 
        \end{tabular}
    \end{center}
\end{table}

In table \ref{tab:generation_time}, two different numbers of graph nodes (50 nodes and 250 nodes represented with the N symbol) along with two texture resolutions (1k and 4k pixels) have been used to show the generation time. Each generation is either a single or multi-layer generation. As seen in the table, the chunks division has a great impact on the generation time. 

\subsection{PLUME rendering}
\begin{figure}[htbp]
\centerline{\includegraphics[width=0.95\linewidth]{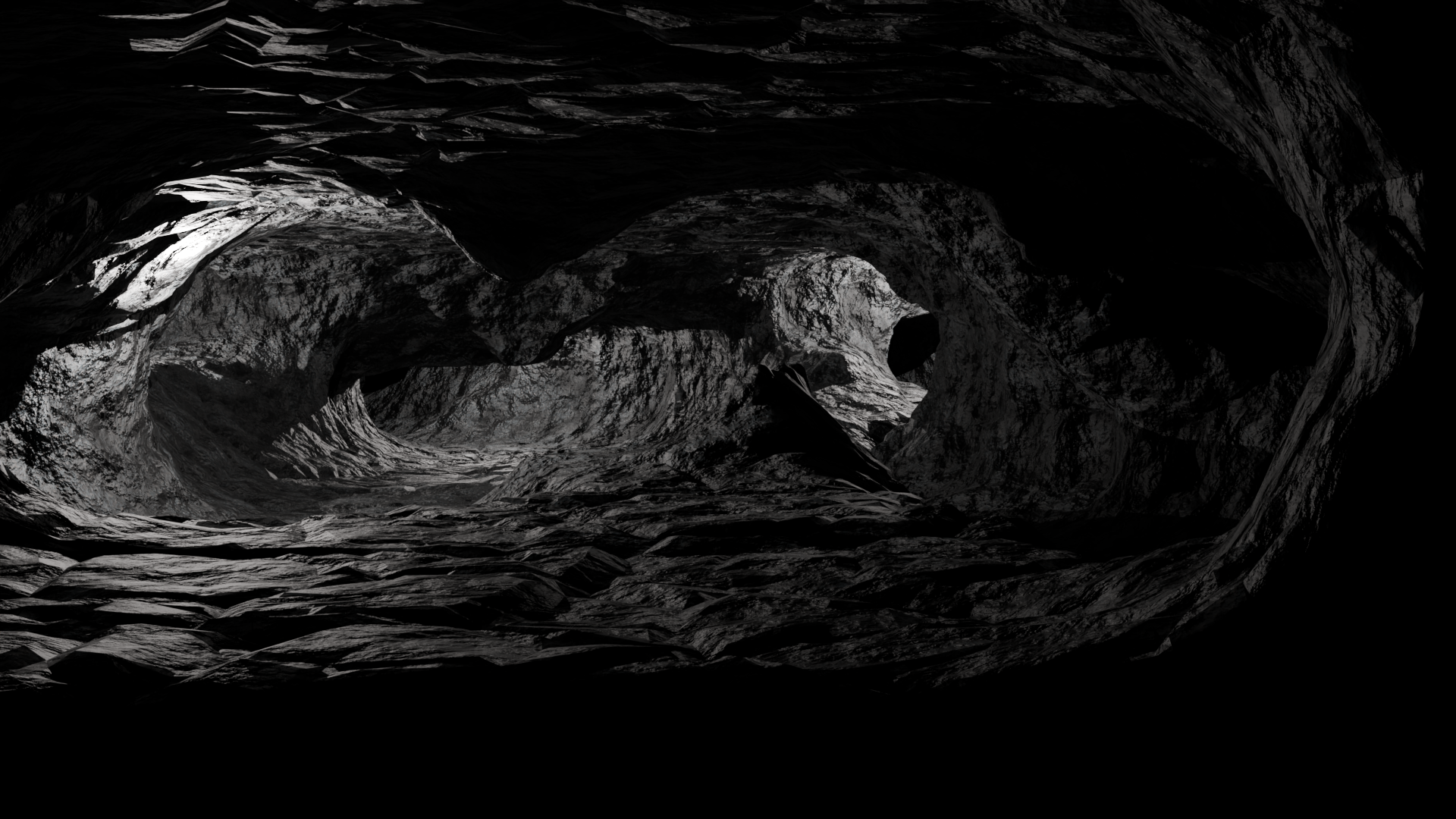}}
\caption{Procedural generated cave with PLUME. Large and open area, low light condition.}
\label{fig:plume_rendering_3}
\end{figure}

Using the previously detailed pipeline, the environments shown in the following Fig. \ref{fig:plume_rendering_1}, \ref{fig:plume_rendering_2}, and \ref{fig:plume_rendering_3} are rendered directly in Blender using Cycles engine. 

\begin{figure}[htbp]
\centerline{\includegraphics[width=0.95\linewidth]{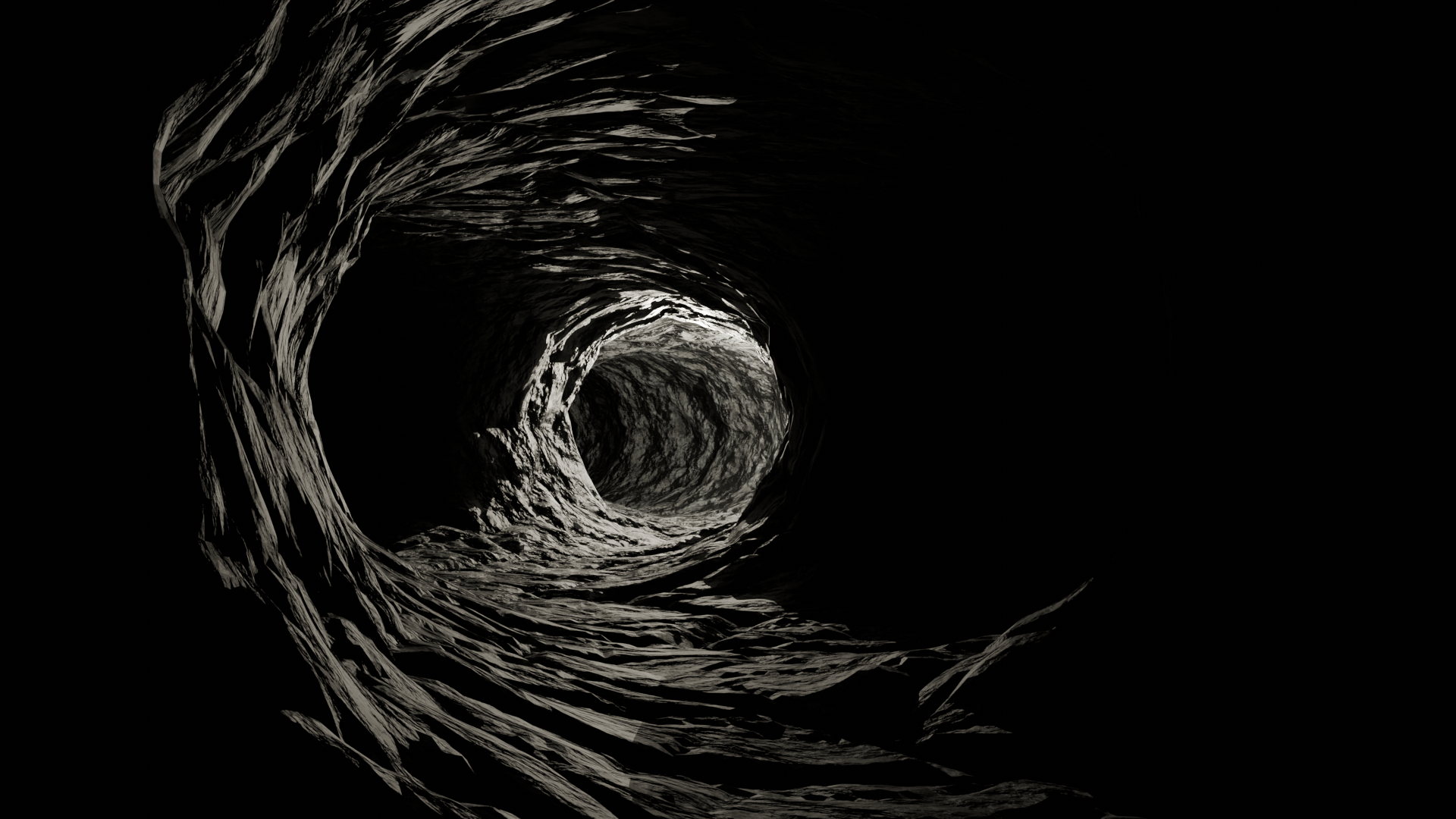}}
\caption{Procedural generated cave with PLUME. High contrast, single light source.}
\label{fig:plume_rendering_1}
\end{figure}

Fig. \ref{fig:plume_rendering_1} shows a challenging environment with a very high contrast between the pitch black area and the central light source. When the rover moves in the environment, the VSLAM algorithm faces significant challenges in detecting strong features. This is because the same area can have a wide luminosity range, making it difficult for the camera to adjust.

\begin{figure}[htbp]
\centerline{\includegraphics[width=0.95\linewidth]{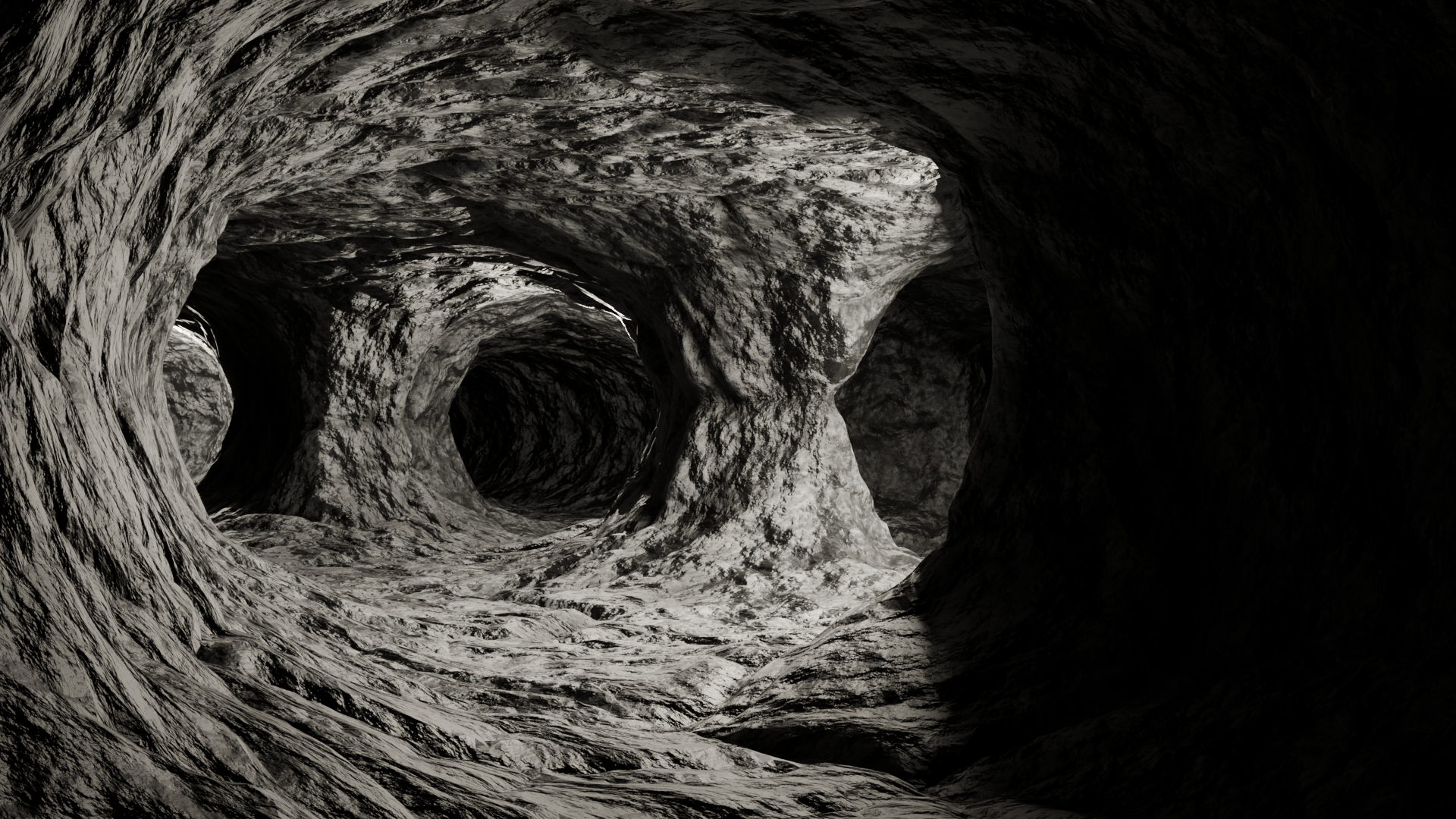}}
\caption{Procedural generated cave with PLUME. Medium contrast, multiple light sources.}
\label{fig:plume_rendering_2}
\end{figure}

On the other hand, Fig. \ref{fig:plume_rendering_2} shows a more open area with a lot of light. The contrast is still as strong as Fig. \ref{fig:plume_rendering_2} however, the illuminated surface is larger simplifying the task for the VSLAM algorithm by providing a larger amount of possible features.

Fig. \ref{fig:plume_rendering_3} displays a complete environment, picturing openings in the passages to the lower layers of the cave, as well as a large central room. This central room shows a different set from the previous two figures, with various elevations and central segmentation along the room. This environment is challenging both for the VSLAM algorithm (as the contrast is still very high) and the physical robot, as it has to traverse a more challenging environment. However, due to the variety of shapes in this generation, the VSLAM algorithm might get a larger amount of strong features compared to the other caves presented before.

\subsection{In simulation}

\begin{figure}[htbp]
\centerline{\includegraphics[width=0.95\linewidth]{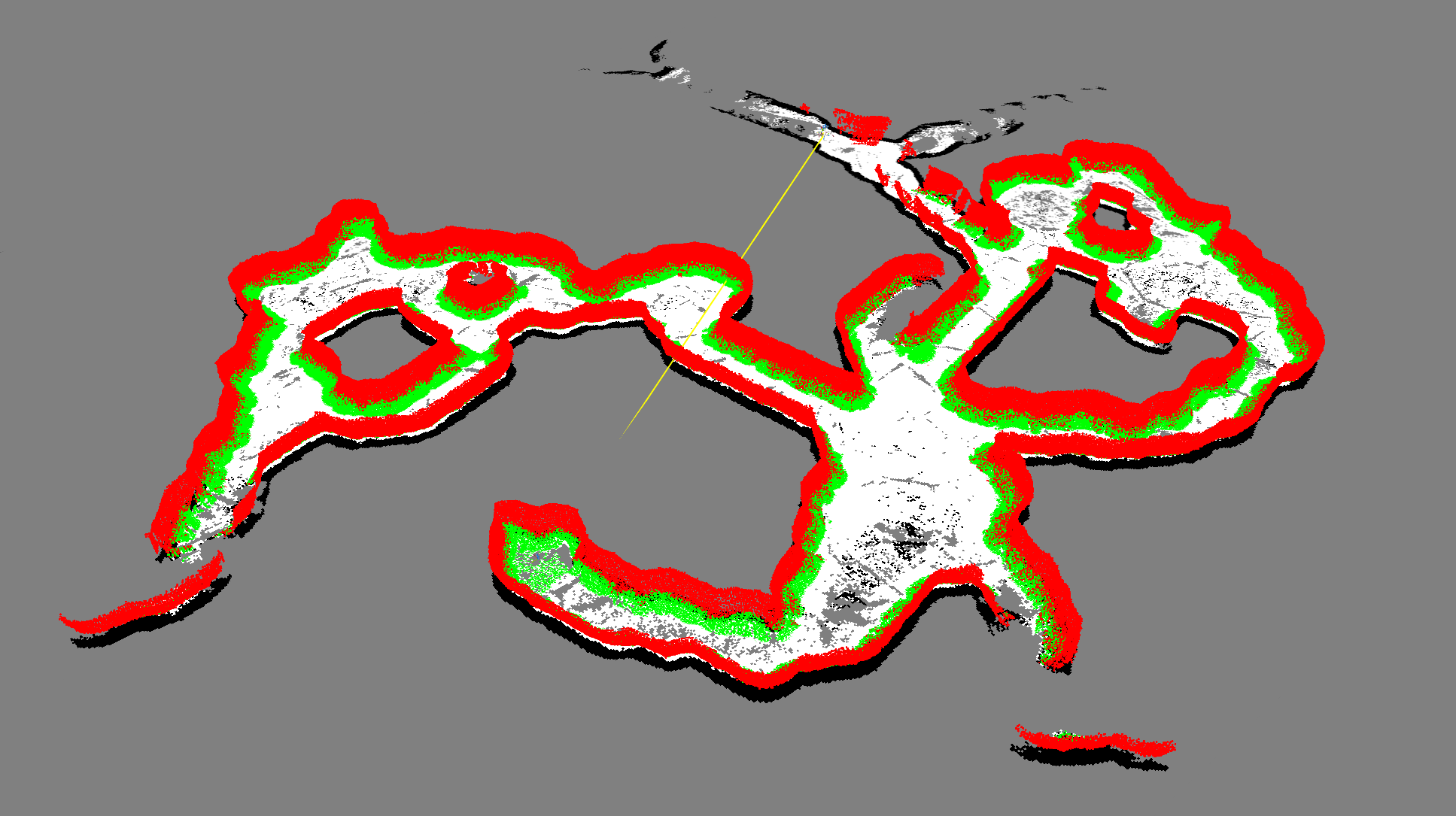}}
\caption{Cloud map (in red and green) as well as 2D map (in black and white) representation in the 3D panel of Foxglove GUI.}
\label{fig:plume_gazebo_1}
\end{figure}

The main purpose of PLUME is to generate procedural environments that are used in a simulator. To this extent, multiple simulations have been conducted using Gazebo to validate exploration, navigation, or traversability algorithms using environments generated with PLUME. The Fig. \ref{fig:plume_gazebo_1} shows the cloud map (in red and green) as well as the 2D map (in black and white) displayed in the Foxglove GUI 3D panel \cite{c21}. This is the result of a 2-hour experiment in the simulator.

During the simulated experiment, a Leo Rover \cite{c20} with an average speed of $0.1m/s$ was used, as well as a 3D LiDAR. The roof of the cave has not been mapped.
The SLAM algorithm used in this specific experiment is RTAB-Map \cite{c19}.

\section{Limitations and future work}
Currently, PLUME has some limitations. PLUME aims to generate environment for wheeled robots, and therefore the environment is designed to be traversable for such a locomotion system. Of course, given a different set of parameters, the environment can be more challenging. However, PLUME only generates terrain without any objects. Rocks or other obstacles can be added to the scene manually, further improving the realism of the cave. In addition, the environment is designed around a graph, and more complex shapes can be included to increase the difficulty of the underground. To such an extent, a group of heterogeneous and autonomous multi-robot systems can be deployed, each type of robot accomplishing a different task of an overall mission in autonomy. Another benefit of the graph structure is that it is very easy to modify the skeleton of the cave by changing the algorithm generating the environment. A possibility is to generate graphs based on real physical simulation instead of procedural generation. This would allow for a more complex and realistic environment, but would largely increase the computation time of the graph.

Future work includes the addition of pyroducts, which are a particular structure of caves formed by an underground lava river. When the lava flow stops, the lava drains downslope and leaves a partially empty conduit, leaving a tube-like shape that needs to be explored. Those environments might be more easily found in space as they can be formed by old volcanoes or asteroid impacts leaving seas of molten rocks and they don't rely on water to be formed. Once the three main underground structures have been added, more studies are needed to evaluate the performance of PLUME to generate a realistic underground environment. Those studies can rely on real data from underground environments to evaluate PLUME.
A final extensions to PLUME would be the multi-layer texture, providing additional information about the physics of the terrain. That information can then be used in a simulation software to add more sticky or slippery areas in the environment, enhancing the overall simulation.

\section{Conclusion}

PLUME is a straightforward, efficient, and modular framework for creating cave-like underground environments relying on a graph to make the skeleton and Blender to generate the mesh and the textures. The generated environments show a high level of realism and detail, suitable for robotic applications. This framework has enabled our team to simulate environments for testing exploration, navigation, and SLAM (Simultaneous Localisation and Mapping) algorithms prior to conducting real-world field experiments. You can find more information on the project here: \href{https://github.com/Gabryss/P.L.U.M.E}{https://github.com/Gabryss/P.L.U.M.E}.




\end{document}